# Coronary Artery Segmentation from Intravascular Optical Coherence Tomography Using Deep Capsules

Arjun Balaji,[a] Lachlan J. Kelsey,[a,b] Kamran Majeed,[c,d,e] Carl J. Schultz,[c,d] Barry J. Doyle[a,b,f,g]

a. Vascular Engineering Laboratory, Harry Perkins Institute of Medical Research, QEII Medical Centre, Nedlands and Centre for Medical Research, The University of Western Australia, Perth, Australia.
b. School of Engineering, The University of Western Australia, Perth, Australia.
c. Department of Cardiology, Royal Perth Hospital, Perth, Australia.
d. School of Medicine, The University of Western Australia, Perth, Australia.
e. University of Ottawa Heart Institute, Ottawa, Ontario, Canada.
f. Australian Research Council Centre for Personalised Therapeutics Technologies, Australia.
g. British Heart Foundation Centre for Cardiovascular Science, The University of Edinburgh, UK.

**Corresponding Author:**
Barry Doyle, Tel: +61 8 6151 1084, Fax: +61 8 6151 1084, Email: barry.doyle@uwa.edu.au
6 Verdun Street, Nedlands, Perth, WA 6009, Australia.
ORCID ID: 0000-0003-4923-2796

**Funding:**

We would like to thank the Western Australia Department of Health Merit Award and the Royal Perth Hospital Medical Research Foundation.

# ABSTRACT

The segmentation and analysis of coronary arteries from intravascular optical coherence tomography (IVOCT) is an important aspect of diagnosing and managing coronary artery disease. Current image processing methods are hindered by the time needed to generate expert-labelled datasets and the potential for bias during the analysis. Therefore, automated, robust, unbiased and timely geometry extraction from IVOCT, using image processing, would be beneficial to clinicians. With clinical application in mind, we aim to develop a model with a small memory footprint that is fast at inference time without sacrificing segmentation quality. Using a large IVOCT dataset of 12,011 expert-labelled images from 22 patients, we construct a new deep learning method based on capsules which automatically produces lumen segmentations. Our dataset contains images with both blood and light artefacts (22.8%), as well as metallic (23.1%) and bioresorbable stents (2.5%). We split the dataset into a training (70%), validation (20%) and test (10%) set and rigorously investigate design variations with respect to upsampling regimes and input selection. We show that our developments lead to a model, DeepCap, that is on par with state-of-the-art machine learning methods in terms of segmentation quality and robustness, while using as little as 12% of the parameters. This enables DeepCap to have per image inference times up to 70% faster on GPU and up to 95% faster on CPU compared to other state-of-the-art models. DeepCap is a robust automated segmentation tool that can aid clinicians to extract unbiased geometrical data from IVOCT.

1. **INTRODUCTION**

Intravascular optical coherence tomography (IVOCT) is a contemporary high-resolution imaging tool used in the assessment of coronary artery disease [1, 2]. It can reveal the arterial lumen geometry and dimensions, as well as vessel wall structure to a limited depth, with near microscopic features (see Figure 1 for examples). Arterial geometry extracted from IVOCT can elucidate information about the luminal area, diameter and wall thickness, all of which are clinically relevant factors that inform the management of disease.

The production and analysis of these geometries has attracted attention, particularly in the automation of geometry extraction [1, 3-5]. Early methods to produce these geometries were semi-automated, requiring software to produce lumen segmentations [3, 6]. However, these semi-automated methods are time and resource intensive and suffer from a range of reproducibility issues that are only discovered when analysis is being conducted [6]. This hinders the use of IVOCT analysis in time-pressured clinical situations.

Recent advancements in machine learning have stirred interest in new approaches to automated lumen segmentation. Since 2015 there have been several publications investigating the use of supervised machine learning techniques such as support vector machines and least squares regression to segment IVOCT data [7-10]. These efforts had varying levels of success but were mainly limited due to the scarcity of training data and the intrinsic features present in coronary artery IVOCT, namely the guide wire's shadow and bifurcations of the coronary artery [7-9]. Newer efforts have used convolutional neural networks and linear regression techniques to annotate lumens using a points method, as opposed to pixel-wise segmentation which have been viable [11].

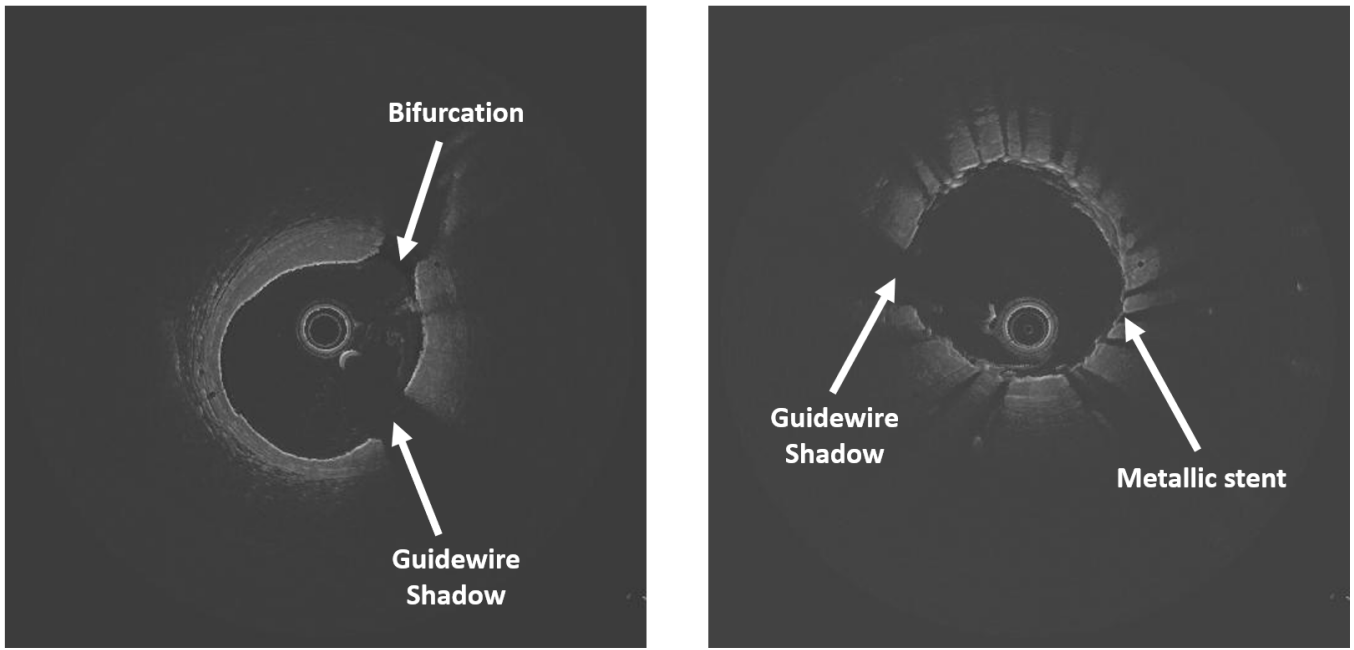

*Figure 1: Two examples of B-scan IVOCT images that highlight some of the artefacts present that make lumen extraction difficult.*

More recently, deep learning methods have demonstrated excellent results at image segmentation in several domains [5, 12]. Deep learning exploits the use of back-propagation, an algorithmic implementation of the chain rule, to iteratively minimize a pre-defined loss function – usually cross-entropy loss [5, 12-18]. As we iteratively alter the internal state of the model to minimize the loss, we can arrive at a model which can map an input image to a desired label, in this case a label of the lumen. Deep learning architectures have garnered a lot of interest and as such there are several state-of-the-art models in literature. Three examples of state-of-the-art models are, in order from oldest to newest: fully convolutional networks (FCN), UNet and DeepLabV3 (DLV3) [12, 15, 17]. Each of these models are designed to be modular and contain an encoder that is based on varying depths of Residual Network (ResNet), where the deeper encoders make the model larger and more computationally intensive to train and run [18].

Many of these state-of-the-art deep learning models tend to be resource intensive during training and inference. For clinical use, we would like the resource burden at inference time to be small. LaLonde and Bagci [13] proposed CapsSeg, a capsule network [14] modified for image segmentation of lung tissue [19]. Capsule networks differ from convolutional networks in that they encode features in vectors instead of scalars, that is, each point in a feature map is occupied by a vector of values instead of just one value [13, 14]. Furthermore, connections between layers are updated using an agreement based routing algorithm in which capsules in a layer (child) are routed to capsules in the next layer (parent) based on the degree to which the child capsule can predict the output of the parent capsule [13, 14]. This means that the routing between capsules is encouraged to converge fast as the routing weights will rapidly grow for related capsules and shrink for unrelated ones. An important introduction was the locally constrained routing algorithm which only routes capsules in a layer to capsules in their neighborhood in the next layer [13]. Usually smaller or shallower neural networks trade-off computational demand for accuracy, but CapsSeg demonstrated that segmentation based on capsules can provide state-of-the-art results in a smaller memory footprint [13]. This inspired us to build an improved, light-weight segmentation model based on capsules, with the aim of reducing computational requirements at inference. This would enable faster segmentation in a clinical setting and lower cost to run on a cloud service. The cost of cloud GPU nodes are significantly more expensive than CPU nodes, so a model that can segment IVOCT data in a clinically suitable timeframe on CPU alone would be cost effective. An IVOCT segmentation algorithm that can run on a mobile CPU/APU would also make new service delivery methods possible in the future.

In addition to the clinical need for automated anatomical data, recent developments in biomechanical engineering have demonstrated the potential of computational fluid dynamics (CFD) to compute the shear stress acting on the endothelium reconstructed from IVOCT [20]. Therefore, if attempting to automate the segmentation of IVOCT, it is sensible to create a framework that can enable CFD as well as create anatomical data. Robust CFD methods require smooth and continuous walls, which are difficult to obtain from IVOCT because of image artefacts that obfuscate the lumen. As IVOCT only images one artery at a time (the guidewire can only inspect a single artery) we would also like our model to handle images with bifurcations but only segment the parent vessel, and also segment stent struts (see Figure 1) which can be difficult [4, 7, 21, 22].

In this work we apply the idea of capsules to coronary artery IVOCT lumen segmentation. With clinical application in mind we created a model with a small memory footprint that has fast inference time while maintaining segmentation quality. In Section 2 we discuss the dataset and the model architecture, in Section 3 we present and discuss our findings and in Section 4 we present our conclusions.

## 2. METHODS

In this work we apply a novel deep learning architecture that involves capsules to coronary artery IVOCT pullbacks which allows us to produce pixel-wise binary masks of coronary artery lumens. We refer to this algorithm as DeepCap. We investigate the effectiveness of two different upsampling regimes, propose a new paradigm for model input selection in an ablation study, and present a fully trained IVOCT segmentation model based on capsules.

### 2.1 Dataset

The dataset we used to train our models was developed in house using IVOCT B-scans acquired as part of the MOTIVATOR Study (ACTRN:12615001234505). All images were acquired using the frequency domain OCT system (C7-XRTM OCT Intravascular Imaging System. St. Jude Medical, St. Paul, MN, USA). Briefly, a 2.7-Fr OCT imaging catheter (Dragonfly, Lightlab Imaging, Westford, MA, USA) was advanced over a standard 0.014 guide wire with the imaging marker sufficiently distal. Automated OCT pullback was performed using a speed of 20 mm/s during simultaneous iso-osmolar X-ray contrast medium (Visipaque 320, GE Healthcare, Buckinghamshire, U.K.) delivery through the coronary guide catheter. All the IVOCT images were anonymized and manually segmented offline using in-house software, where images were first segmented by one user and then verified by another user; these segmented images represent the binary ground-truth label. The dataset of 12,011 images, from 22 patients, contains blood and light artefacts (22.8% of images), as well as metallic (23.1%) and bioresorbable stents (2.7%). The size and quality of this dataset, with artefacts, enables the formation of a robust and reliable machine learning workflow. Once the dataset was compiled in its entirety, we randomly split it into three distinct sets; a training set (70% of total dataset), a validation set (20%) and a test set (10%), with pullbacks randomly assigned to each data subset, while ensuring the size ratio. The reason cases are grouped together is that they contain images similar to each other, so having one of these images in the training set and another in the test set would not be fair. The purpose of a validation set is to perform "mini-tests" per epoch to continuously monitor the progress of our network in an objective fashion. It is important to remember that the model is prohibited from learning when being run on the validation or test set. Transformations are not applied to the validation or test set, giving a more accurate representation of real-world use.

### 2.2 Preprocessing

We feed DeepCap an input that comprises of some combination of three distinct images; an augmented input image, a 2-dimensional (2D) Gaussian derivative of the input image and an axial forward and backward difference image.

We transform the raw image data (360 × 720 pixels) that has been extracted by the imaging system software into Cartesian form before taking a central crop of 300 × 300 pixels which has the effect of removing black bars from the borders of the image. The reasoning behind using Cartesian form is based on published literature showing that there is no statistical difference between Polar and Cartesian form in IVOCT vascular bifurcation classification tasks [23]. Furthermore, we found that IVOCT images are easily interpretable when in Cartesian form and neither the images nor masks are distorted. The model can be easily adapted to perform segmentation on Polar images. This Cartesian 300 × 300 pixel image represents a single *pre-augmented* input image. It has been shown [24] that augmenting the training dataset of a deep learning model can improve its generalizability for image segmentation tasks and thus we apply a sequence of transforms on our *input* before presenting it to the model for training. These transforms are; a random crop of 256 × 256 pixels, a horizontal reflection, a vertical reflection, a Gaussian blur ($\alpha = 1$), a clockwise/counterclockwise rotation ($0° \leq \theta \leq 360°$), and salt and pepper noise [24]. The random crop is always applied, while the other transformations are applied with a probability of 0.5 each. The model receives a completely new perspective of the training data every epoch due to the probabilistic application of these transforms, which improves the generality of the final model. We use the same random seed between experiments to force deterministic application of the transforms.

Secondly, we compute a 2D-Gaussian derivative of the input image which computes a combination of band-pass filtering and spatial derivatives [25]. This is useful in the coronary artery IVOCT imaging domain because there is a scarcity of trainable data for the model, thus by providing a supplementary structure that removes background noise and provides some heuristic edge detection we can speed up training of the model and extract more low-level features from the training data.

Finally, we compute a forward and backward difference about the input image in the axial plane which can be expressed as,

$$\Delta X(z_0) = X(z_0 + \delta) - X(z_0 - \delta), \qquad (1)$$

where $X(z)$ is the image array at $z$ and $\delta$ is the distance between slices as reported by the IVOCT software. We believe this will allow the model to consider the behavior of the images in the locality of the image being segmented before making a prediction about the lumen geometry.

We performed experiments, presented in Section 3.2, to investigate the model's performance when given an input of either just the input image, the input image and the 2D-Gaussian derivative, the input image and the axial difference image and finally all three images.

## 2.3 The Model Architecture

The model's architecture, as illustrated in Figure 2, is reminiscent of the UNet's with its downsampling branch, upsampling branch and skip connections however there are several key differences. The first layer of the model is a conventional convolution operation that converts the input to a set of four feature maps each of which is a 64 × 64 grid of 16 dimensional vectors. These vectors are known as the primary capsules. The convolution kernels that produce the primary capsules have randomly initialized weights and this is the only layer of the network that will not undergo dynamic routing [13, 14]. We will denote the shape of a set of capsules as $(M, H, W, D)$ where $M$ is the number of feature maps, $H$ and $W$ are the height and width of a feature maps respectively and $D$ is the dimension of the capsule vectors.

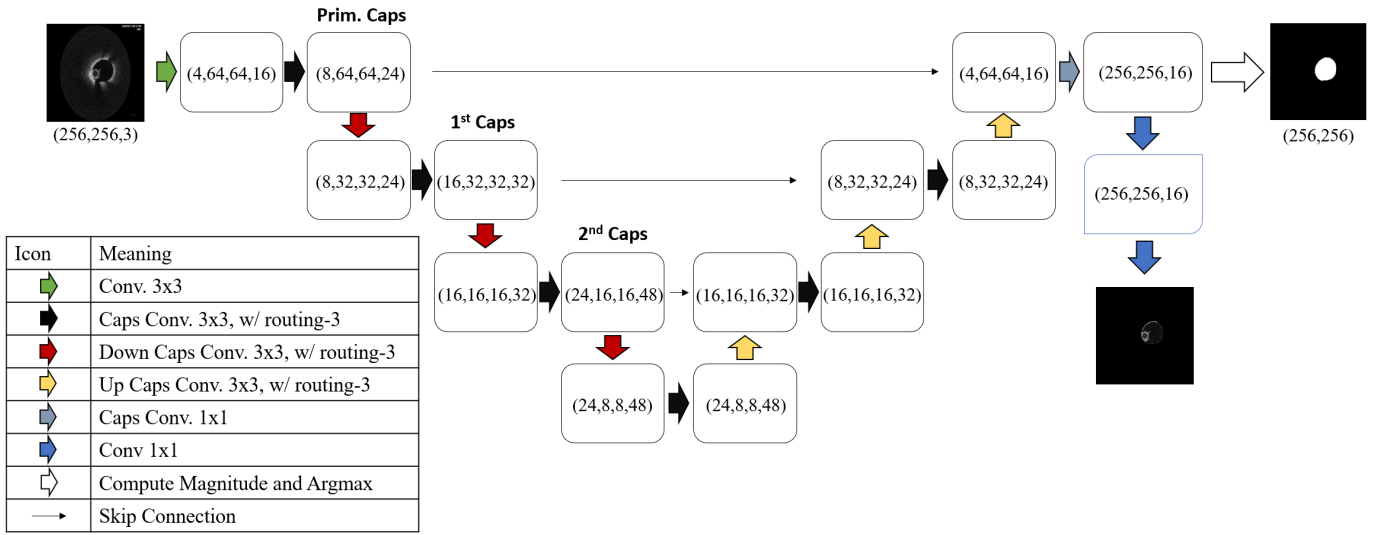

*Figure 2: A schematic of the model's architecture that highlights how data flows through the various layers of our network. The legend explains the name of the layer, the kernel size (if applicable) and the amount of routing iterations (if applicable). Prim. (Primary), 1st and 2nd capsules layers are labelled as such.*

The 'down' portion of the network consists of the 3 layers, each of which is comprised of two separate capsule convolution modules and their associated routing schemes. The first of these modules takes in an input of shape $(M^l, H^l, W^l, D^l)$ and outputs a tensor of shape $(M^{l'}, H^l, W^l, D^{l'})$, and the second of these modules takes in that output of shape $(M^{l'}, H^l, W^l, D^{l'})$ and outputs a tensor to layer $l+1$ of shape $(M^{l'}, H^{l''}, W^{l''}, D^{l'})$ where $M^{l'}, D^{l'}, H^{l''}$ and $W^{l''}$ are specified at each module. The values in our experiment are given in Figure 2. It is useful to briefly describe the mechanics of the operations here, though they are detailed in [13]. In the first convolutional capsule module, we convolve the input tensor with a $k^{l'} \times k^{l'}$ set of learned, nonlinear transformation matrices which produces $M^{l'}$ prediction vectors ($\hat{\boldsymbol{u}}_{xy}$) that are $D^{l'}$-dimensional for each space on the $H^l \times W^l$ grid. To arrive at the input to a capsule we compute the weighted-sum over the prediction vectors where the weights ($c_{l|xy}$), known as routing weights, are determined by the routing algorithm specified in Sabour et al. (2017) We can express this operation as follows,

$$\boldsymbol{p}_{xy} = \sum_n c_{l|xy}\, \hat{\boldsymbol{u}}_{xy}$$

where,

$$c_{l|xy} = \frac{\exp(b_{l|xy})}{\sum_k \exp(b_{l|k})}$$

and $b_{l|xy}$ are the log prior probabilities that $\hat{\boldsymbol{u}}_{xy}$ should be routed to the capsule $\boldsymbol{p}_{xy}$. The final output capsule ($v_{xy}$) is then computed using the non-linear squashing function as expressed in Sabour et al. (2017),

$$v_{xy} = \frac{\|p_{xy}\|^2}{1 + \|p_{xy}\|^2} \frac{p_{xy}}{\|p_{xy}\|}.$$

As illustrated in Figure 2 we used $k^{l'} = 3$ which means that each set of prediction vectors sent to the next module has information about the eight neighboring capsules. This dynamic, locally constrained routing algorithm is how the model's capsule weights are optimized, while the convolutional kernel weights are optimized by backpropagation. This is a key difference to regular convolutional networks which use backpropagation as the optimization scheme for all model weights. The second module works similarly but instead of altering the number of feature maps or the dimension of the capsule vectors, we employ a stride in our convolution to downsample the feature maps. However, no information is wholly discarded during the downsampling because the kernel size of $k^{l''}$ is larger than the stride size (in Figure 2, note that $k^{l''} = 3$ but the $stride = 2$). At the end of each module we perform 3 iterations of the locally constrained dynamic routing algorithm as prescribed in [13].

The 'up' portion of the network is characterized by two properties; upsampling of current feature maps and skip-connections that combine higher-level extracted features with lower-level extracted features. We have a choice of upsampling methods at our disposal and here we will compare transposed convolutional upsampling with bilinear interpolation. The motivation behind using bilinear interpolation to upsample feature maps is that others [13, 14] show how capsules encode information about the thickness, pose and spatial orientation of objects in the image, it therefore would be reasonable to simply scale these feature maps to the size required for skip-connections. The use of transposed convolutions as an upsampling method is motivated by the strong results they have demonstrated in prior convolutional neural networks [13-15]. The upsampling methods are integrated into a convolutional capsule layer and are followed by three iterations of the locally constrained dynamic routing algorithm. After the upsampling module we employ our previously utilized convolutional capsule layers again to increase the depth of our network and allow it to extract more complex feature relationships from the training data.

After the input has gone through the 'down' and 'up' portions of our network we use two consecutive convolutional capsule layers to produce an output tensor of shape $2 \times 256 \times 256$. We use a Gaussian blur layer, with a kernel size of 3 and a sigma of 2, here to ensure that any checkboard artifacts are removed from the final output. We would like the 1$^{st}$ channel of the 0$^{th}$ dimension to contain the probability that the pixel belongs to the background class and the 2$^{nd}$ channel to contain the probability that the pixel belongs to the lumen class. We impose this by applying a softmax function over the 0$^{th}$ dimension of the tensor.

During training we compute a soft Dice score (SDS) and a binary cross entropy loss with the model's prediction and the target label, which can be expressed as,

$$\mathcal{L}(p, y) = -(y \ln(p) - \ln(1 - p)) + \lambda \frac{2\,py}{p + y}$$

where $p$ is the prediction from the model, $y$ is the target label and $\lambda$ is a factor to scale the contribution of the SDS to the total loss. We use a $\lambda$ of 0.05 in our experiments to ensure the binary cross entropy dominates the total loss, which is required to ensure the loss can reliably converge. These losses were chosen because our model's output map is an array of probabilities and our desired target is a binary array. Our choice of these two training losses is also grounded in literature as they have demonstrated state-of-the-art results in varied image segmentation tasks, particularly in the medical imaging domain [13, 15, 26, 27]. The SDS is a popular measure of similarity between the output masks from the model and ground-truth labels, in our experiments it quantifies how well the model can segment the IVOCT lumen in a given image which is an easily interpretable metric. We apply an argmax function over the 0$^{th}$ dimension of the prediction to arrive at the final binary output.

## 2.4 Experiments

Our model was built on *PyTorch* and we used an Adam optimizer with a cosine annealed, one-cycle learning rate scheduler with a peak learning rate of 0.001 and a 90% split between warmup and cooldown as per the one cycle learning rate scheduler outlined in Smith (2018) [28]. We used GPU acceleration on a *Nvidia P100* to train our network with a batch size of 24. All preliminary experiments were run on 10 epochs. Throughout the analysis of our model's results, we focus on six key metrics; the SDS, pixel sensitivity, pixel specificity, Hausdorff distance, number of parameters and time taken for inference on a batch. As explained in the introduction, a potential use of this model is to construct geometries for CFD simulations and so we are interested in the predicted labels relative to the actual labels. This is different to conventional geometrical analysis where clinicians are interested in the absolute area of predicted labels. SDS is a meaningful proxy of absolute luminal area and we demonstrate this by showing a case later (Section 3.4, Figure 8), where absolute luminal area and radius are compared on a per image basis between a human made label and one made by DeepCap. Pixel sensitivity and specificity allow us to investigate how well the model can classify pixels as lumen or background separately. Hausdorff distance is used in conjunction with the above metrics as it is not affected by vessel size. Larger vessels have better SDS because the proportion not segmented by the model is small, but this is not the case for the Hausdorff distance [29]. The number of trainable parameters in a model gives us an indication of its complexity and memory footprint which is important to our investigation as we are aiming to have this model run in a cost-effective manner in the cloud or on a local device in a clinic that may have limited access to a GPU. Finally, we include the time-taken to run inference on an image as this is will indicate effectiveness in time-sensitive clinical situations. The reason behind choosing these five metrics to quantify the model's success is that they provide the clearest understanding of how the model performs in terms of segmentation quality and inference properties. The Jaccard Index and F1 score are all related intrinsically to the soft Dice score, sensitivity and specificity.

Our first experiment was to investigate the model's performance under the bilinear interpolation and transposed convolution upsampling regimes. For these experiments both models were run on the same dataset split to eliminate any sample selection bias. The input selection was all three inputs. Once we decided which upsampling to use we performed an ablation study to investigate which combination of inputs yields the best model performance. The four experiments were: (i) only the input image, (ii) the input image and the 2D-Gaussian derivative, (iii) the input image and the axial difference map, and (iv) all three. Because the input image is present in all four experiments, we will henceforth term these experiments; IM, 2DG, ADM and ALL respectively. Again, the dataset split was the same for these experiments.

We compared our model to a UNet-ResNet18 (UNet-18), FCN-ResNet50 (FCN-50) and DeepLabV3-ResNet50 (DLV3-50) to compare our model to other state-of-the-art segmentation algorithms. UNet was introduced in 2015 and has received considerable research focus since then [15, 30, 31]. UNet-18 is a UNet with an 18-layer residual network as an encoder [15, 18] while FCN-50 and DLV3-50 have 50-layer residual networks as their encoder [12, 17, 18]. To compare how quickly these models converge, we train them both on the same data split for 30 epochs and then evaluate them on the same hold test set.

Finally, we present a DeepCap that was trained until the validation loss stagnated for 10 epochs and evaluate it against other published coronary artery IVOCT segmentation methods.

## 3. RESULTS AND DISCUSSION

We performed several experiments to investigate the effectiveness of our model in different scenarios and with different metrics, which are explained below.

### 3.1 Upsampling Paradigms

An important decision to make about our proposed model is the type of upsampling technique to use. We investigated using bilinear interpolation and transposed convolution to upsample feature maps in the second part of the network. Besides the upsampling method, both experiments used the exact same parameters - for example, the number of layers and kernel sizes. The inference batch size was 48 images in both experiments and as such the inference time is the time taken to process a batch divided by the batch size of 48. The results are shown in Table 1.

*Table 1: Performance of the model in the bilinear interpolation and transposed convolution upsampling schemes.*

| Upsampling method | Transposed Convolution | Bilinear |
|---|---|---|
| soft Dice score ($\mu \pm \sigma$) | **96.25 ± 5.73** | 95.06 ± 7.46 |
| Median | **97.76** | 97.60 |
| Min-Max | 48.05-99.30 | 48.35-99.20 |
| Sensitivity ($\mu \pm \sigma$) | **93.27 ± 8.22** | 91.35 ± 10.90 |
| Median | **95.59** | 95.31 |
| Min-Max | **34.22-98.61** | 31.88-98.41 |
| Specificity ($\mu \pm \sigma$) | 99.54 ± 0.75 | 99.41 ± 0.90 |
| Median | 99.72 | 99.70 |
| Min-Max | 90.67-99.95 | 92.40-99.93 |
| Inference time (ms/image) | 39 | 40 |

From Table 1 we can see that upsampling via transposed convolution is more effective than bilinear interpolation at maximizing SDS, sensitivity and specificity after 10 epochs of training. As the data is paired but not normally distributed, we use a paired Wilcoxon signed-rank and find that there is a statistically significant difference between using transposed convolution over bilinear interpolation as an upsampling method (p<0.001). Transposed convolutional upsampling achieved a 1.3% better mean SDS and a 23.2% smaller standard deviation in SDS in a holdout test set. Mean pixel sensitivity was 2.1% better in the transposed upsampling regime than in bilinear interpolation and this corresponded with a 24.5% smaller standard deviation. Pixel specificity in the test set was similar between the two upsampling regimes with no clear winner. Inference time was the same for each model with a batch of 48 images being processed in ~1.9 seconds equating to a single image being done every 40ms.

## 3.2 Determining Optimal Inputs

We propose adding 2D-Gaussian derivatives and axial difference maps as additional inputs to the model to increase segmentation accuracy. To test this, we performed four experiments to identify which combination of the three proposed inputs (IVOCT image must be selected) would result in the best lumen segmentations. Like earlier, these experiments were conducted with the same dataset split and all model and environment parameters were kept constant, except the input data.

Table 2 shows us that providing all three images as inputs to the model gives us the best segmentation results on a holdout test set after 10 epochs of training. Using a paired Wilxocon signed-rank test we find a statistically significant difference between using ALL over IM, 2DG and ADM ($p<1.0\times10^{-60}$, $p<1.0\times10^{-30}$, $p<1.0\times10^{-36}$, respectively). ALL had a 4% better mean SDS than IM and a 28% smaller standard deviation which means that inclusion of the local context features we proposed does lead to better and more consistent segmentations. Improvements in pixel sensitivity are even greater as ALL has a 7.5% higher mean than IM and a 29% smaller standard deviation. This means that lumen pixels are more effectively classified as such when we provide the model with extra inputs that contain information about the lumen either side of that slice. The ALL model only contains 0.4% more parameters than the IM model implying that we receive a disproportionate gain in SDS for the increase in parameters (4% increase in mean SDS for 0.4% more parameters).

Table 2: A table summarizing the performance of 4 different input image schemes on soft Dice scores, pixel sensitivity and specificity on a holdout test set. Batch size of 24, transposed convolutional upsampling and 10 training epochs use in each case. I (Image), G (2D-Gaussian Gradient), A (Axial Difference). Data presented as mean ± standard deviation.

| Inputs Selected | I (*IM*) | I + G (*2DG*) | I + A (*ADM*) | I + G + A (*ALL*) |
|---|---|---|---|---|
| Soft Dice score (μ±σ) | 92.39 ± 7.98 | 93.48 ± 9.20 | 93.04 ± 10.04 | 97.54 ± 2.75 |
| Median | 95.57 | 96.77 | 96.61 | 97.76 |
| Min-Max | 52.36-99.10 | 23.67-99.07 | 15.57-99.03 | 48.05-99.30 |
| Sensitivity (μ±σ) | 86.73 ± 11.89 | 88.88 ± 13.01 | 88.25 ± 13.60 | 95.31 ± 4.37 |
| Median | 91.53 | 93.75 | 93.45 | 95.62 |
| Min-Max | 35.46-98.21 | 13.43-98.16 | 8.44-98.08 | 31.62-98.60 |
| Specificity (μ±σ) | 99.11 ± 0.91 | 99.20 ± 1.44 | 98.91 ± 2.62 | 99.69 ± 0.39 |
| Median | 99.45 | 99.59 | 99.59 | 99.72 |
| Min-Max | 93.08-99.86 | 81.74-99.89 | 69.67-99.92 | 89.39-99.93 |
| No. of Parameters | 4,984,496 | 4,985,296 | 4,985,296 | 4,986,240 |

We see modest improvements in the means of test set metrics and a deterioration of standard deviations when comparing 2DG and ADM to IM. This shows that having extra context in only one of either the image plane or axial direction hinders the model's performance, which is also supported by Figure 3. However, when we provide context in both the image plane and the axial direction, we see significant improvement in both means and standard deviations of all three metrics. We believe this is because image scenarios that benefit from extra context in the axial direction are independent from ones that benefit from extra context in the image plane. For example, context in the axial direction would be useful when we are approaching a bifurcation but extra context in the image plane would be more useful when a blood artefact is obscuring the lumen border. It is possible for these two scenarios to occur in the same region, but it is not always the case. Therefore, when only one extra contextual input is provided, the model maximises the contribution of that input in scenarios when it is useful. However, when it is not useful, it can minimise it, but not totally get rid of it, thus increasing the variability in test set metrics. Though, when we provide both pieces of extra contextual information the model can discriminately alter the contributions of those extra inputs depending on the image features it extracts.

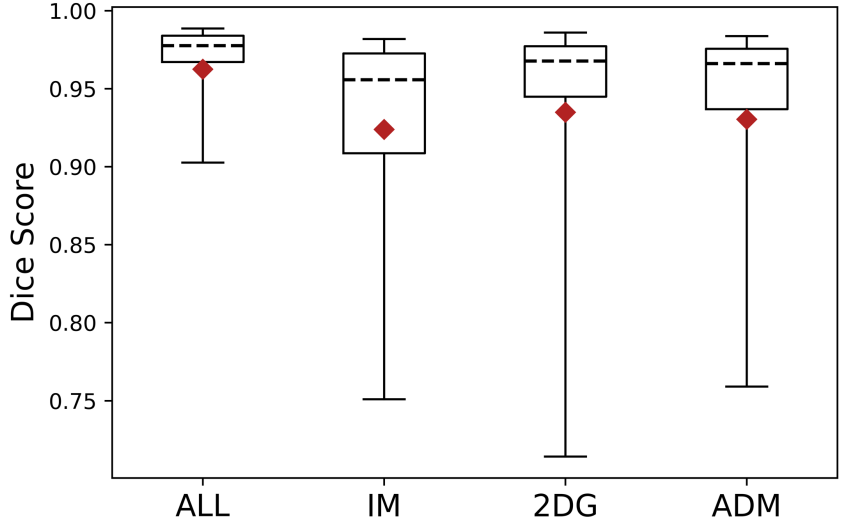

*Figure 3: A series of box and whisker plots illustrating the 95$^{th}$-percentile bound of soft Dice scores for each input regime. The dotted line marks the median and the red diamond marks the mean of the samples. Note that the y-axis is truncated at SDS = 0.70.*

### 3.4 Comparison to state-of-the-art models

From our prior experiments we found that transposed convolution upsampling and providing the model with all three proposed inputs yielded the best lumen segmentations, so we trained our complete model with these properties. Table 3 compares our model's performance to UNet-18, FCN-50 and DLV3-50 on a holdout test set after 30 epochs of training each.

Table 3 illustrates that DeepCap outperforms UNet-18 on our hold out test set with respect to SDS, Hausdorff distance and pixel sensitivity but not pixel specificity. Using a paired Wilcoxon signed-rank test we see a statistically significant difference in SDS and Hausdorff distance between the two models ($p<0.0005$, $p<0.0001$, respectively). DeepCap has a 2% higher mean SDS and a 21% smaller standard deviation of SDS than UNet-18 which implies that it segments lumens with higher quality and more consistently. Both FCN-50 ($p=0.02$) and DVL3-50 ($p=0.007$) have marginally better SDS than DeepCap. DeepCap performs better than DL3 with respect to the Hausdorff distance ($p=0.003$) but not better than FCN ($p=0.098$). On GPU, DeepCap is 29% faster than UNet-18, 60% faster than FCN-50, and 72% faster than DLV3-50. On CPU, DeepCap is 39% faster than UNet-18, and takes a negligible amount of time compared to FCN-50 and DLV3-50.

*Table 3: Results of DeepCap vs UNet-ResNet18 on a holdout test set after training for 30 epochs with batch size 16. The DeepCap SDS distribution is significantly different to UNet-18, FCN-50 and DLV3-50 with p-values of 0.0005, 0.02 and 0.007 respectively.*

| Metrics | DeepCap | UNet-ResNet18 | FCN-ResNet50 | DeepLabV3-ResNet50 |
|---|---|---|---|---|
| Soft Dice score ($\mu \pm \sigma$) | 97.00 ± 5.82 | 95.06 ± 7.46 | 97.39 ± 3.51 | 97.54 ± 2.75 |
| Median | 98.29 | 98.01 | 98.11 | 98.03 |
| Min-Max | 48.05-99.30 | 48.35-99.20 | 60.17-99.26 | 59.80-99.30 |
| Hausdorff Dist. ($\mu \pm \sigma$) | 3.30 ± 1.51 | 3.57 ± 1.77 | 3.33 ± 1.35 | 3.32 ± 1.25 |
| Median | 2.96 | 3.14 | 3.10 | 3.14 |
| Min-Max | 1.43-16.85 | 1.68-16.27 | 1.58-15.97 | 1.62-16.03 |
| Sensitivity ($\mu \pm \sigma$) | 93.27 ± 8.22 | 91.35 ± 10.90 | 95.09 ± 5.49 | 95.31 ± 4.37 |
| Median | 96.55 | 95.31 | 97.03 | 96.17 |
| Min-Max | 34.22-98.61 | 31.88-98.41 | 67.30-99.40 | 42.66-98.60 |
| Specificity ($\mu \pm \sigma$) | 99.54 ± 0.75 | 99.41 ± 0.90 | 99.73 ± 0.27 | 99.69 ± 0.39 |
| Median | 99.72 | 99.70 | 96.29 | 99.75 |
| Min-Max | 90.67-99.95 | 92.40-99.93 | 43.03-98.52 | 3.60-99.93 |
| Inference time GPU (ms/image) | 4.4 | 6.2 | 10.1 | 15.5 |
| Inference time CPU (ms/image) | 97 | 158 | 2468 | 3343 |
| No. of Parameters | 4,986,096 | 31,113,008 | 32,947,986 | 39,633,986 |
| Disk Size (Megabytes) | 36.0 | 142.3 | 132.1 | 159.0 |

Figure 4A compares the 95% confidence interval between the four models, and we can see that DeepCap performs well despite taking up significantly less computational resources than the other models. Figure 4B also demonstrates that DeepCap can segment arteries to an SDS of higher than 97%. We found that DeepCap performs poorest in scenarios where the input image is atypical or contains more than one artefact (see blue example in Figure 4B and further examples in Supplementary Figures S1-2). During our analysis we found that for images with an SDS greater than 98%, masks were near-perfect to the human eye. In our test set, more than 63% of images segmented by the model had an SDS greater than 98%, meaning in a 200 image B-scan that 126 images on average would not need human supervision. For comparison, the proportion of images segmented better than 98% for the UNet-18, FCN-50 and DLV3-50 were 53%, 56% and 52% respectively. We expect that as we continue to train the model with more difficult cases and more fringe scenarios, this number will reduce further.

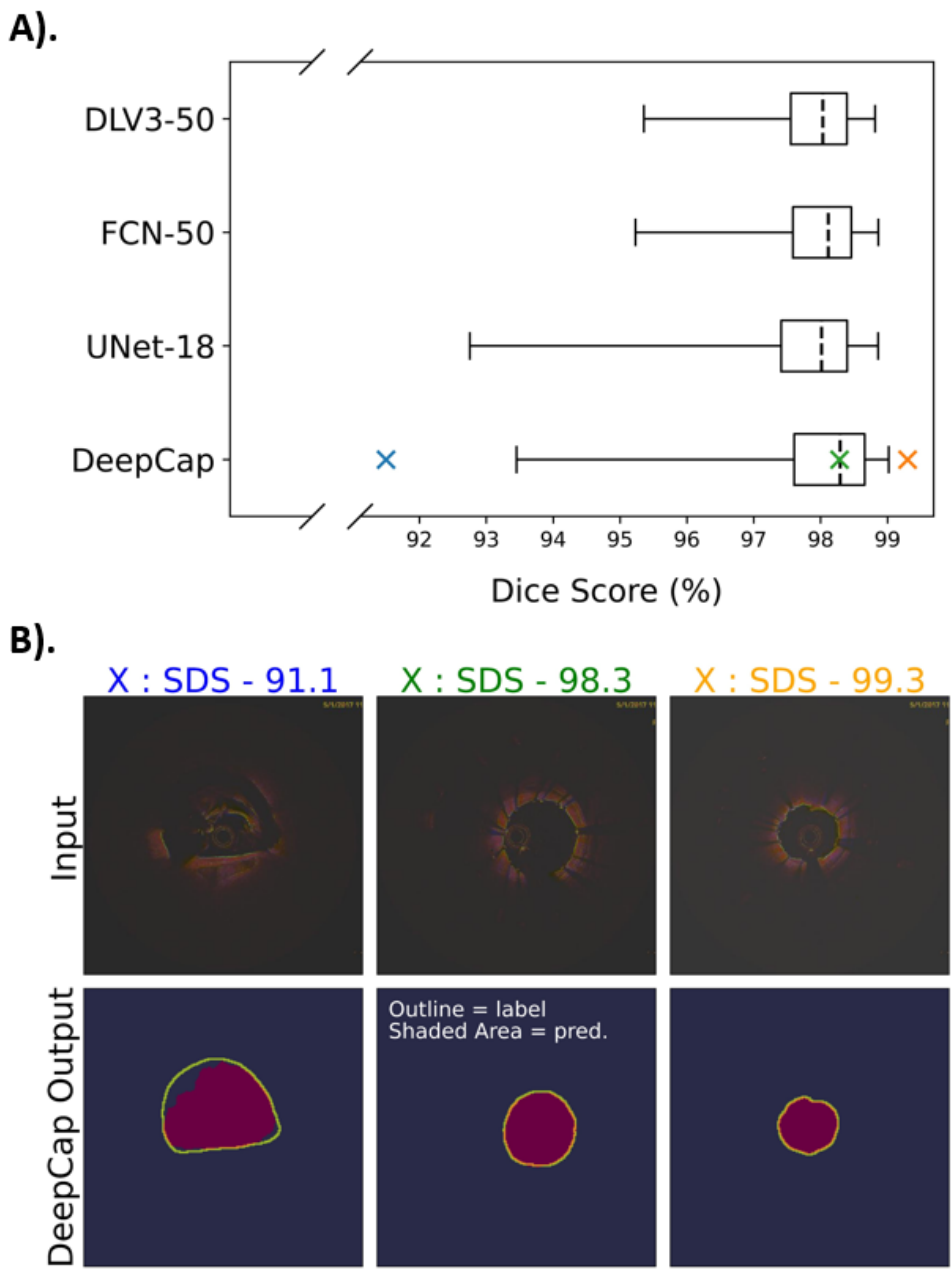

*Figure 4: (A) Median centered 95th percentile comparison between DeepCap and three state-of-the-art models. (B) We show three examples of DeepCap's outputs on test set images (red area) with the ground truth label (yellow outline). These three examples illustrate examples of DeepCap's performance based on SDS: poor (blue), at the median (green), and excellent (orange). Note that the blue example is rare for DeepCap, outside of the 95% confidence interval, in our test set but this artefact is caused by inadequate flushing during image acquisition, which is a common artefact.*

Furthermore, DeepCap achieves these results while taking up less disk approximately a quarter of the disk space as each of the other comparison models. Figure 5 highlights how DeepCap manages to maintain a high SDS while having significantly less parameters compared to the other comparison models.

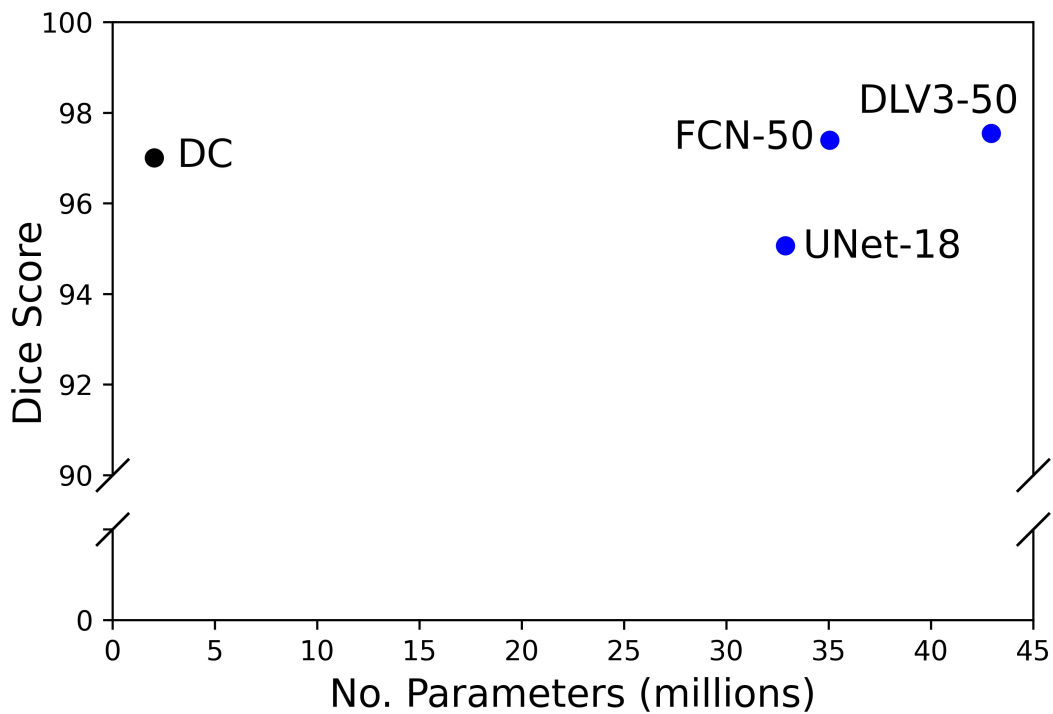

*Figure 5: DeepCap compared against state-of-the-art models with respect to the tradeoff between number of parameters and test SDS. Note the y-axis break.*

Though we have not specifically focused on how well the areas of segmented lumens compare to human ground truth segmentations, Figure 6 demonstrates that the efficacy of our model is maintained over the varying lumen sizes present in the test set. We can see that a majority of the points lie within the 95[th] median centered percentiles of the comparison models, which indicates the model is performing as expected. Importantly, Figure 6 shows that our model is effective over different luminal areas, and to reiterate, DeepCap can achieve this performance by using a fraction of the parameters of the comparison models.

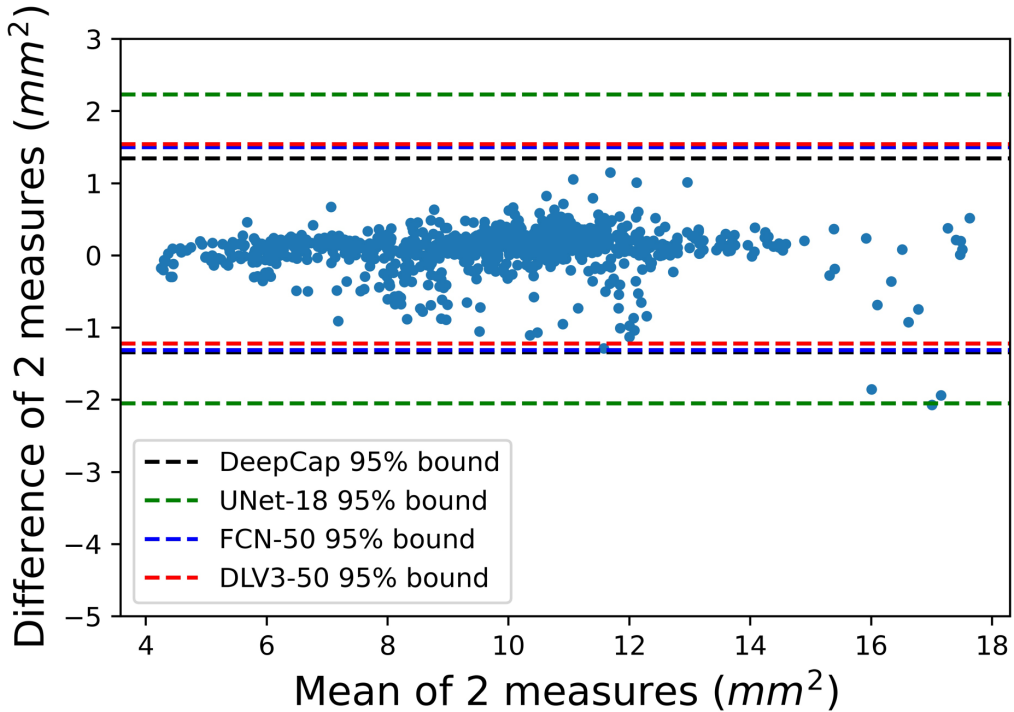

*Figure 6: Bland-Altman plot demonstrating DeepCap performance compared to the human labelled ground-truth on our test set. The 2 measures are DeepCap and human labelled ground-truth. Lines show the 95$^{th}$ median centered percentiles of the four models under investigation.*

### 3.5 Comparison to existing models for intravascular OCT

We continued to train DeepCap for another 30 epochs, for a total of 60 epochs, until the validation loss stagnated for 10 epochs and we present a comparison between this fully trained DeepCap and other literature in the field (Table 4). It should be noted that direct comparisons of research are difficult due to the differences in datasets however the dataset we have used is more than seven times larger than the one used by Miyagawa et al. (2018), 400 times larger than the one used by Kim et al. (2014), and 41 times larger than the one used by Moraes et al. [32-34]. This also does not consider the diversity of data which is pertinent in the context of coronary artery IVOCT lumen segmentation as the presence of artefacts, stents and plaque will all affect the validation performance of the model and its ability to generalise in real world applications. We purposely included all image data, irrespective of artefact and noise, so as to build a robust model that can handle typical IVOCT images encountered in the clinic.

Table 4: Summary of dataset sizes and fully trained model's performance over a single random split of the dataset relative to existing algorithms. Data presented as mean ± standard deviation. N/A are undisclosed data in the publication. * indicates median value.

|  | DeepCap | Miyagawa et al. [34] | Kim et al. [32] | Moraes et al. [33] | Tsantis et al. [35] | Yong et al. [11] |
|---|---|---|---|---|---|---|
| Images (Total) | 12,011 | 1,689 | 30 | 290 | 2710 | 19,027 |
| Training | 8,502 | 1,352 | N/A | N/A | N/A | 13,342 |
| Validation | 2,499 | 337 | N/A | N/A | N/A | 5,685 |
| Test | 1,010 | N/A | N/A | N/A | N/A | N/A |
| Sensitivity (%) | 95.05 ± 6.69 | 96.92 ± 3.75 | 99.21 ± 0.51 | 99.29 ± 2.96 | 91.00 ± 1.00 | N/A |
| Specificity (%) | 99.66 ± 0.56 | 99.11 ± 2.19 | 99.70 ± 0.15 | 96.31 ± 2.88 | 96.00 ± 2.00 | N/A |
| Soft Dice score | 97.31 ± 4.52 | 94.34 ± 11.71 | N/A | N/A | N/A | 98.50* |

A comparison between DeepCap and the model presented by Miyagawa et al. illustrates the performance benefit afforded by a more comprehensive IVOCT dataset and the use of capsules over regular convolution [34]. Metrics of note are DeepCap's higher mean SDS (97.31 vs 94.34) and significantly smaller deviation of SDS (4.52 vs. 11.71). The training time was 3 hours and 40 minutes which indicates that this network architecture can quickly learn the features present in IVOCT inputs, an important property as this could be useful in the future to quickly retrain the model or add new cases to its segmentation repertoire. Due to the size of DeepCap only being 36.0Mb, this model will be capable of running in a timely manner on mobile or CPU in intraoperative or other clinical settings. DeepCap can segment a 200 image B-scan in 0.8 seconds on GPU (*Nvidia P100*) and in 19 seconds on CPU (single core). The minimal computational requirements of DeepCap widens the potential use cases of this model in clinics with limited access to expensive computer hardware.

### 3.6 Fully trained and tested DeepCap

Figure 7 illustrates the robust segmentation potential of our fully trained model. These images demonstrate some of the more difficult segmentation scenarios as they contain more than one image artefact per image. We can see that that our model is achieving near mean and median SDS on some of these images which has two implications; (i) we can expect the model to perform as expected in a clinical context where patients have stented segments and blood artefacts in their IVOCT images, and (ii) masks will not require human supervision as their SDS exceeds 98%.

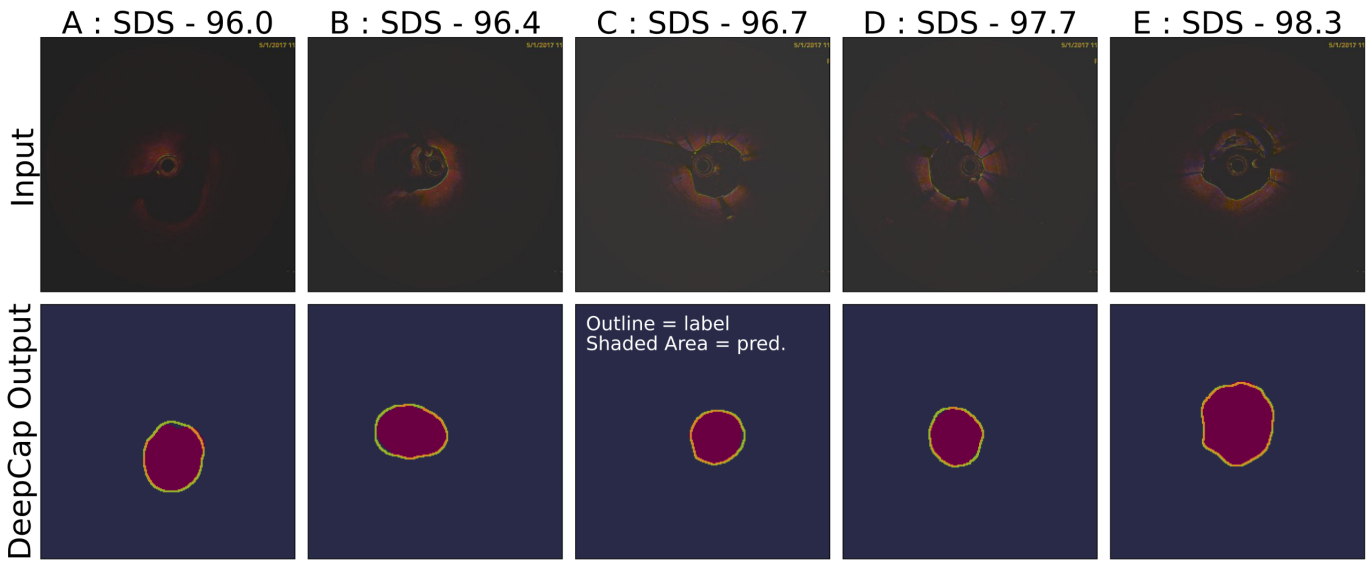

*Figure 7: A sample of four images from the validation set that demonstrate the model's robust segmentation ability. A, B, C and D contain bifurcations and blood artefacts and C, D and E contains stent and blood artefacts. The model's segmentations are depicted by the red image masks and true lumen labels are depicted by the solid yellow outline.*

To demonstrate the type of features that the capsules extract from the input IVOCT images, we include a sample of feature maps from the down portion of the network in Figure 8. We chose an input that the model was effective at segmenting because it more easily elucidates the way the model works. We can see that in the primary capsule layer (Prim. Maps in Figure 8) the model is extracting very low-level features such as the different kind of curves and contrast changes. In the first capsule layer (1st Maps in Figure 8) the model is discerning more high-level image features such as the guidewire shadow and the central imaging source. We can also clearly see that the model recognizes the bounds of the circular IVOCT image. Finally, the second capsule layer (2nd Maps in Figure 8) identifies higher-still image features such as the gross location of the lumen and shadow. An interesting observation about these images is they look like a shadowed sphere; this indicates some of the generality gained by these networks due to the repeated linear transformations we apply to the input signal.

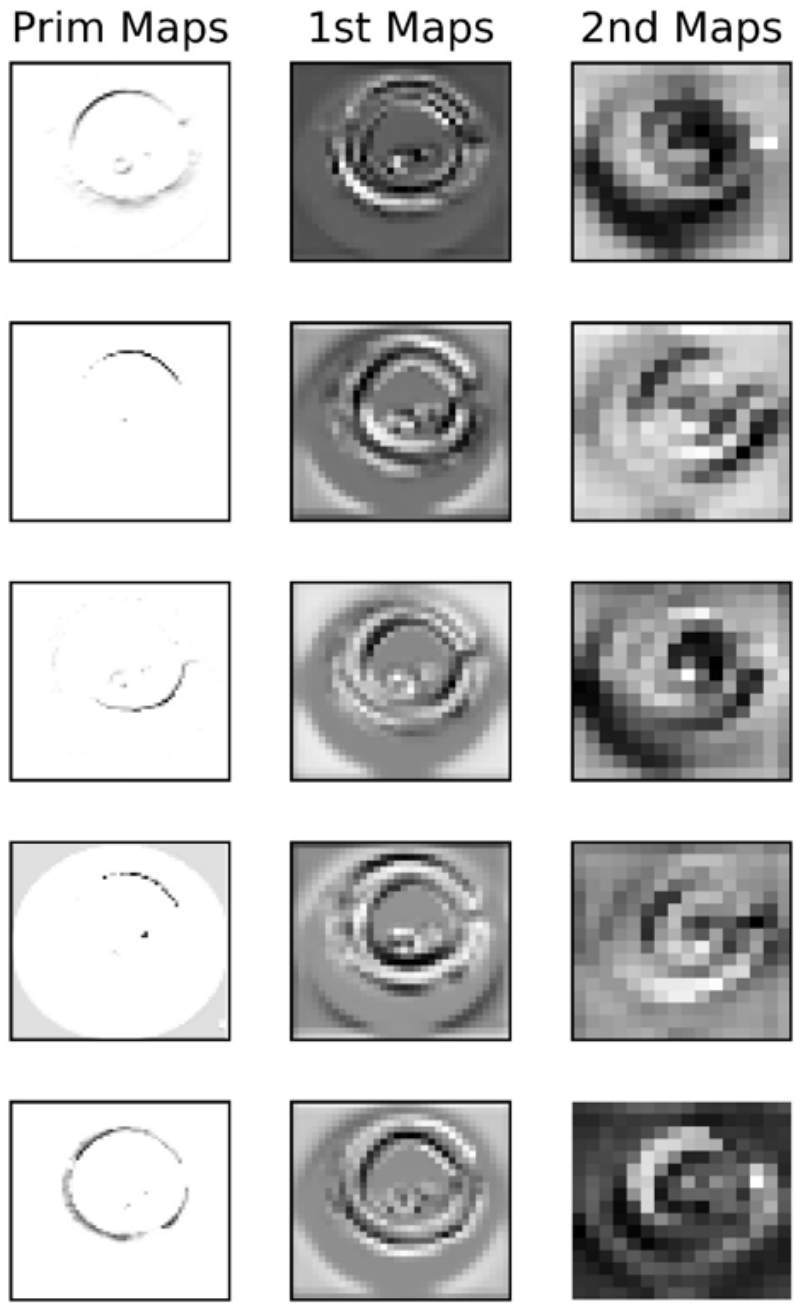

*Figure 8: A sample of extracted feature maps by the model from the down layer of the network. Each column of images is a small subset of the extracted feature maps, the exact number can be found in Figure 2. The input image was the same as Figure 5(1O). Prim = primary capsule layer; 1st maps = first capsule layer; 2nd map = second capsule layer.*

In Figure 9 we present an example case from the validation set that compares our model's segmentation against two human ones; H1 being in the validation set, H2 being performed after the model was finalized and blinded from both the DeepCap prediction and H1. Both human segmentations were performed using an in-house software with manual segmentation tools. The 3D reconstructions used a marching cubes algorithm to convert the voxel-data to a triangular mesh (Figure 9). This case is the only OCT-pullback in the validation set containing images of a bioresorbable stent (51 images) and contains a large bifurcation, making it time-consuming to perform precise manual segmentation (i.e. > 1 hour). The second human segmentation allows us to analyze how our model performs when we account for ground-truth segmentation variability. Furthermore, as only 241 images containing bioresorbable stents were present in the training set (241/9,608), for these images the performance of the model is considered to be limited by the dataset. Figure 8 demonstrates that our model captures global and local characteristics of the human segmentations. We see that our model performs similar to humans when comparing vessel radii on a per-point basis in IVOCT lumen segmentation. The radius heat maps of H1, DeepCap and H2 are similar and show that even in the stented regions, the model can produce segmentations that yield radii measurements comparable to human segmentations. Figure 9 also shows that our model's segmentations yield similar cross-sectional areas to those of human segmentations. In the stented region, the model produces smoother outputs than the human segmentations, resulting in a slightly larger cross-sectional area between images 210 and 250. For each of the three geometries reconstrcuted, we also determined the wall shear stress and pressure drop obtained from computational fluid dynamics simulations (Figure S3 and Table S1) to further demonstrate that the performance of DeepCap is comparable to a human.

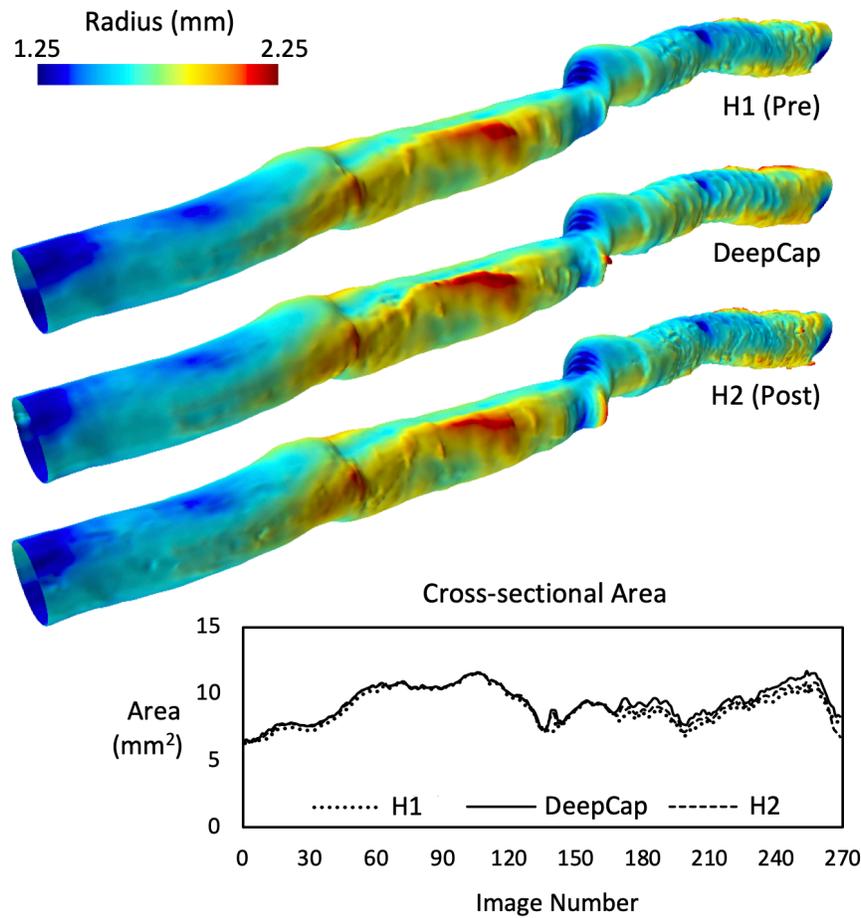

*Figure 9: A comparison of our model (Prediction) vs. two human segmentations (H1, H2) for a case from our validation set. For each segmentation, the radius heat map illustrates the magnitude of displacement between the vessel-wall vertices and the lumen-mask centroids, computed for each image using the area-moment method. The plot compares the cross-sectional area (mm$^2$) per image yielded by the various segmentations. The surfaces shown here are generated using a marching cubes algorithm at a voxel resolution of 100μm × 50μm × 50μm. The radial heat maps are shown on smoothed surface meshes, obtained using STAR CCM+ (v13.06, Siemens): each surface mesh underwent a Laplacian smoothing operation where the vertices were re-projected to the initial surface.*

There is significant scope for future work on automated IVOCT lumen segmentation using deep learning. For instance, our model ignores bifurcations and instead segments the lumen in images with bifurcations based on interpolation from previous frames. Nevertheless, we have demonstrated that capsules form the basis of an effective architecture for IVOCT segmentation because they are able to easily localize image features and relate those image features to output lumen segmentations. One limitation of this model which can be noted in Fig 4A. is that despite our model producing adequate segmentation in a higher percentage of cases than the comparison algorithms the frequency of sub-optimal segmentations is also higher. To solve this we require making our model to be more effective at segmenting fringe cases. Increasing the proportion of the dataset that contains artefacts and bio-resorbable stenting will improve the ability of the model to segment lumens in these fringe cases. In the future we will investigate a new training scheme where more difficult images are fed back into the model at training time so that the model is exposed to these more difficult images at a higher frequency. We are also investigating adding a model feature that predicts the potential accuracy for an input image. This would allow us to create an exclusion criterion and improve usability in a clinical setting.

## 4. CONCLUSIONS

We propose a new deep learning model based on capsules (DeepCap) as an accurate and efficient method of automatically segmenting coronary artery lumens from IVOCT. We designed DeepCap with a clinical use case in mind and as such it does not require image modification or exclusion prior to being able to segment the lumen. We trained our model on one of the largest expert-labelled coronary artery IVOCT datasets in the medical machine learning literature. In our analysis we investigated several different design schemes for the internal upsampling regime of the model and model input selection. We found that our model performed the best, as measured by mean pixel sensitivity, specificity and soft Dice score, when we used the transposed convolutional upsampling scheme and gave the model the image, a corresponding 2D-Gaussian derivative and an axial difference map, as inputs. Finally, we compared the fully trained DeepCap against three state-of-the-art segmentation algorithms; a UNet-ResNet18, FCN-ResNet50 and a DeepLabV3-ResNet50 architecture, and showed that DeepCap achieves competing metric scores on our test set and is faster at inference time and takes less disk space. Our model demonstrates the potential for capsules as a basis for IVOCT lumen segmentation models and we show that providing the model with local spatial information does enhance performance. Furthermore, DeepCap is fast, being able to segment an entire 200 image B-scan in about 0.8 seconds on GPU and 19 seconds on CPU only. Compared to the existing methods, our novel model of automated coronary segmentation will provide the critical luminal information faster, during both time-dependent percutaneous coronary intervention procedures and offline image analysis.

**Model Performance Compared to State-of-the-Art**

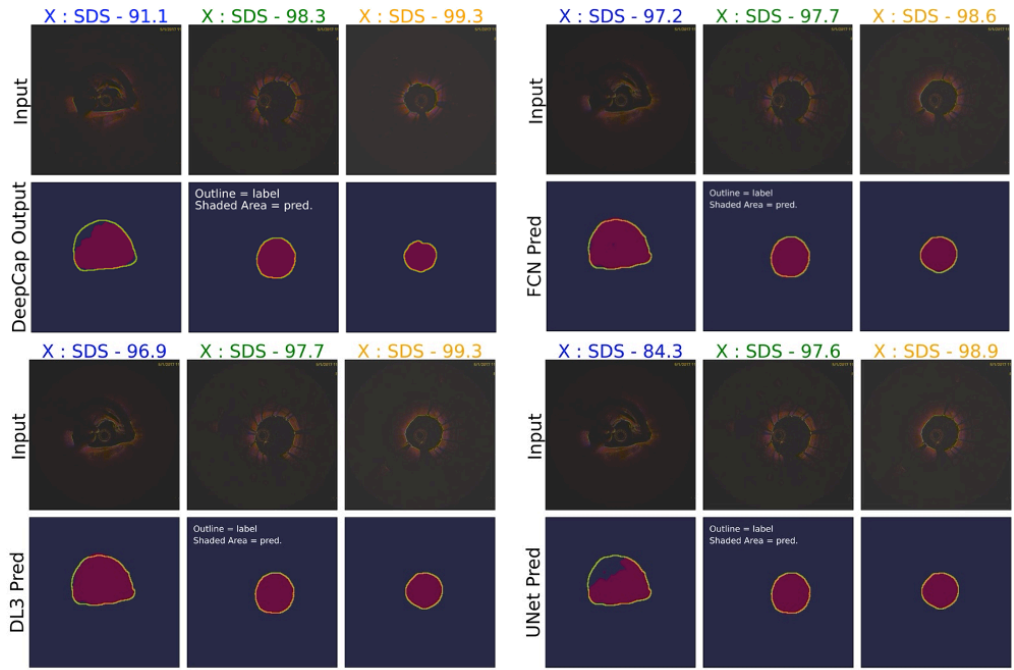

Figure S1: Comparison of DeepCap outputs (top left) to the other models tested. This figure corresponds to Fig. 4B in the main paper.

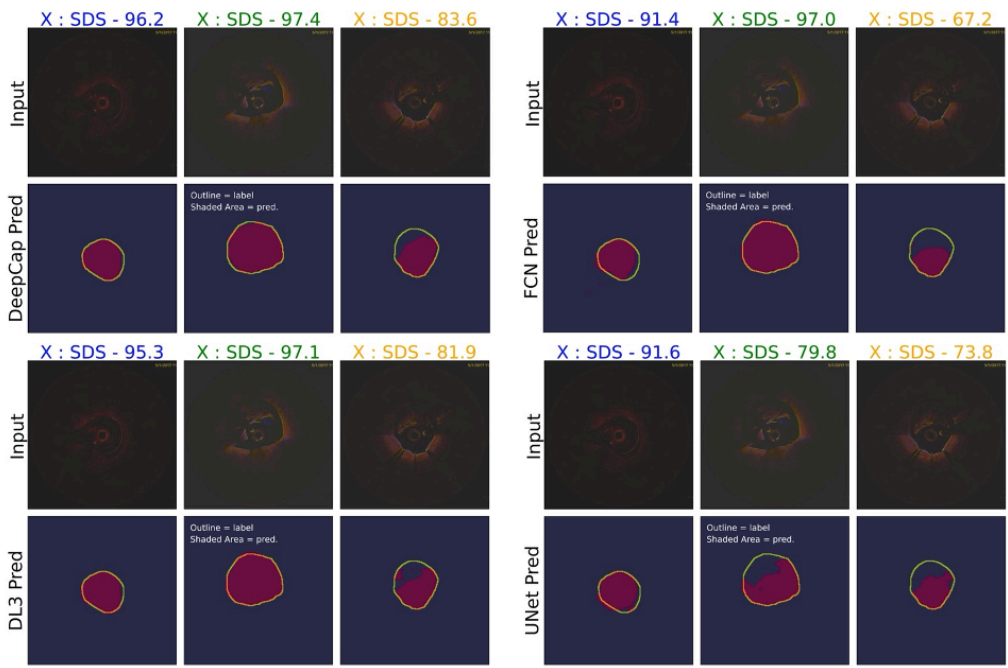

Figure S2: Comparison of DeepCap outputs (top left) to the other models tested. This figure illustrates the performance of the models on images that are typically difficult to segment due to poor quality images and presence of stents. DeepCap outperforms all three existing models.

**Reconstructed Geometries for Computational Fluid Dynamics**

The single conduit CFD models were constructed and solved using a hexahedral finite-volume discretisation and segregated flow solver in STAR-CCM+, with anisotropic boundary layer refinement used to capture near-wall velocity gradients. The total number of cells/elements in each of the three models were 951K (Pre-Label), 960K (DeepCap Prediction) and 1160K (Post). The material, blood, was defined as having a density of 1050 kg-m$^{-3}$ and a non-Newtonian shear thinning viscosity, following the Carreau-Yasuda model. The steady-state simulation was performed to determine relevant hemodynamic parameters. An arbitrary inlet flow rate was set that achieved average wall shear stress (WSS) values in the physiological range.

The models are lacking branches and are not registered as the registration processes requires an additional imaging modality, such as coronary computed tomography or angiography, to provide the vessel path in three-dimensions, as well surrounding branches: necessary to provide accurate boundary conditions. Hence, these results only show the influence of the local geometric changes on relevant haemodynamic quantities in the OCT image-space.

Table S1 shows that compared to the human segmentation data, DeepCap's prediction obtains similar WSS and pressure drop ($\Delta P$). Furthermore, despite good qualitative agreement (Figure S3), the prediction provides a smoother representation of the geometry and hemodynamic quantities in the stented region, where minor local features were not captured. However, it is expected that this may be overcome if more images of bioresorbable stents were present in the training set (2.7%). Nevertheless, the deep learning segmentation method is considered to be fit for the development of hemodynamic models that measure quantities such as shear stress and pressure drop. Note, for the same boundary conditions, material properties and numerical discretization, the WSS and pressure drop results are dependent on both surface features and the vessel diameter ($\propto d^a; a > 2$).

*Table S1: A summary of the difference in observed Wall Shear Stress (WSS) as measured in Pascals (Pa) and Pressure Drop ($\Delta P$) as measured in mmHg between geometries extracted from the Pre-Label, DeepCap Prediction and Post-Label.*

| Geometry | $WSS_{avg}$ ($Pa$) | $\Delta P$ ($mmHg$) |
|---|---|---|
| Label | 1.82 | 1.19 |
| Post | 1.73 (-5.0%) | 1.14 (-4.2%) |
| DeepCap | 1.70 (-6.5%,-1.7%) | 1.10 (-7.5%,-3.5%) |

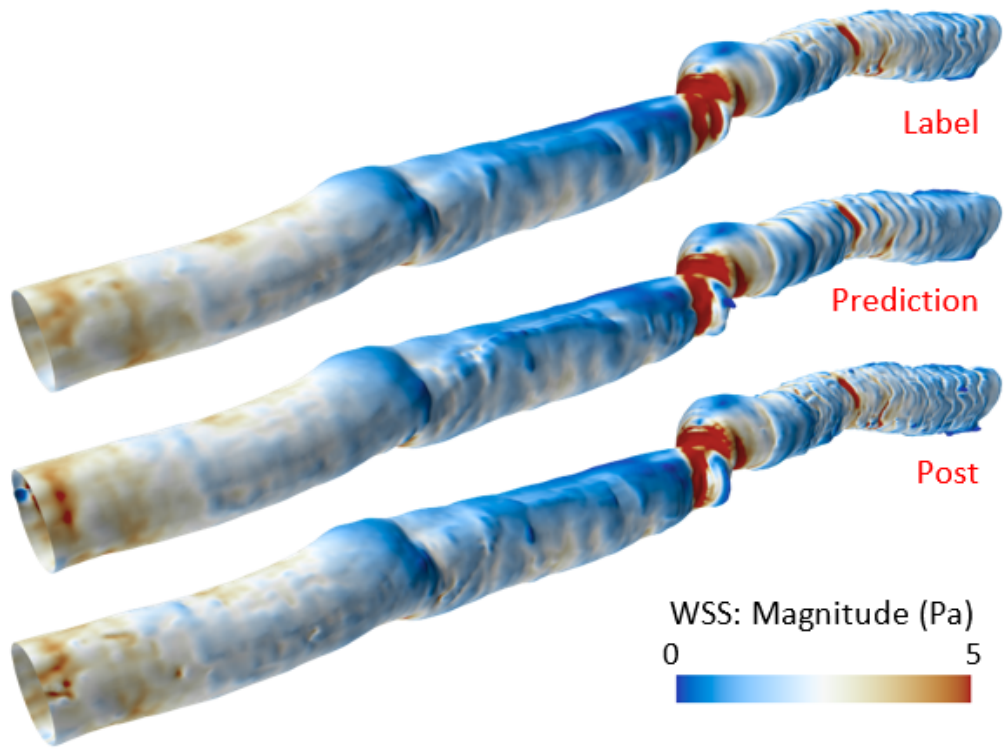

*Figure S3: Wall Shear Stress (Pa) contours for the three geometries; Pre-Label (Label), DeepCap's Prediction (Prediction) and Post-Label (Post). Note the color scale in the bottom right corner.*